\documentclass[letterpaper, 10 pt, conference]{ieeeconf} 

\IEEEoverridecommandlockouts

\usepackage{graphicx} 
\usepackage{amsmath}
\usepackage{amssymb}

\usepackage{hyperref}
\usepackage{color}

\usepackage{tikz}
\usetikzlibrary{shapes.geometric, arrows.meta, positioning, fit, backgrounds, calc}

\title{\LARGE \bf

Continuous Online Fault Detection for Mobile Robots via Adaptive Edge Models

}

\author{Jordan Levy$^{1,2}$, Nicolas Verstaevel$^{1}$, Vincent Talon$^{2}$ and Benoit Gaudou$^{1}$% <-this % stops a space
\thanks{*This work was supported by the company Soben and the ANRT}% <-this % stops a space
\thanks{$^{1}$IRIT, Université Toulouse Capitole, Toulouse, France}%
\thanks{$^{2}$TwinswHeel, Soben, France}
}

\begin{document}

\maketitle
\thispagestyle{empty}
\pagestyle{empty}

% \tableofcontents

\begin{abstract}
	Mobile robots require robust, real-time fault detection capable of continuous adaptation on constrained edge hardware. While deep time-series models excel at unsupervised anomaly detection, their computational cost prohibits high-frequency onboard execution. This paper bridges this gap via a Teacher-Student distillation framework. An offline foundation model (TSPulse) generates pseudo-labels from unlabeled time series augmented with fault injections. A lightweight MiniRocket Student, adapted with a Recursive Least Squares estimator, approximates this complex decision boundary to execute real-time inference onboard. Evaluations on the TSB-AD benchmark and a physical mobile robot demonstrate the Student achieves a 4.30 ms CPU inference latency. During real-world domain shifts, online adaptation enables the Student to recover from unseen mechanical degradation, improving VUS-PR scores from 0.26 to 0.75 without catastrophic forgetting. Crucially, an uncertainty-guided active learning strategy minimizes operator cognitive load, requesting sparse interventions only when encountering novel fault distributions. These results validate the deployment of state-of-the-art anomaly detection on resource-constrained robotics through offline-to-online distillation.
\end{abstract}

\section{Introduction}

As mobile robots are increasingly deployed in complex, hazardous, and unpredictable environments, ensuring their operational reliability has become a critical challenge~\cite{khalastchi2018fault, delmerico2019current}. Despite advances in hardware and control architectures, robots remain susceptible to mechanical degradation, sensor failures, and unexpected external disturbances. Consequently, robust Fault Detection (FD) systems are essential to prevent catastrophic failures, minimize downtime, and ensure safe human-robot interaction~\cite{khalastchi2018fault, isermann2005fault}. Typically, upon detecting a fault, the robot can transition into a safe degraded mode (e.g., reducing its speed) and alert a distant human operator who remotely supervises the fleet. This human-robot interaction point provides an opportunity not only for immediate recovery, but for continuous online adaptation.

Traditionally, FD relied on physics-based models requiring exhaustive, system-specific knowledge of the robot's kinematics and dynamics. However, the rising complexity of modern robotics makes deriving and maintaining these exact models challenging~\cite{khalastchi2018fault}. To overcome this limitation, the robotics community has shown a surging interest in data-driven FD~\cite{amato2025data}. Data-driven FD bypasses the need for comprehensive prior physical modeling by leveraging available sensor data, excelling at uncovering hidden, non-linear correlations across multidimensional sensor arrays~\cite{park2018multimodal}.

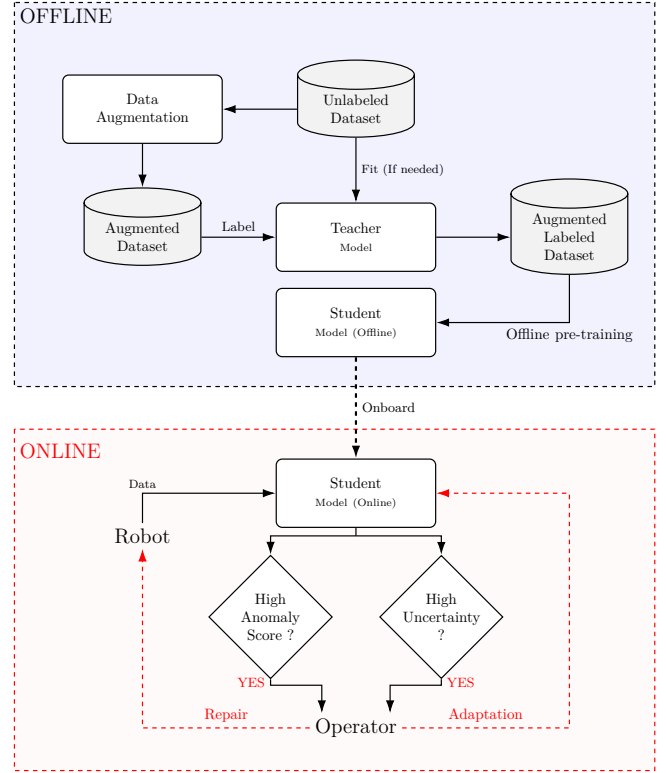
\begin{figure}[htb!]
	\resizebox{\columnwidth}{!}{
		\begin{tikzpicture}[
			>={Latex[length=3mm, width=2mm]},
			database/.style={
					cylinder, draw, thick, shape border rotate=90,
					aspect=0.25, text width=2.5cm, align=center,
					minimum height=1.5cm, fill=gray!10
				},
			block/.style={
					rectangle, draw, thick, rounded corners,
					fill=white, text width=3.5cm, align=center,
					minimum height=1.6cm
				},
			decision/.style={
					diamond, draw, thick, fill=white,
					text width=1.8cm, align=center, inner sep=0pt
				},
			entity/.style={
					rectangle, align=center, font=\bfseries
				},
			line/.style={draw, thick, ->},
			redline/.style={draw, thick, dashed, red, ->}
			]

			% ==========================================
			% BACKGROUND BOXES
			% ==========================================
			\draw[draw=black, dashed, thick, fill=blue!5] (-7, 9) rectangle (8, 0);
			\node[below right, font=\Large\bfseries] at (-7, 9) {OFFLINE};

			\draw[draw=red, dashed, thick, fill=red!2] (-7, -1) rectangle (8, -9);
			\node[below right, font=\Large\bfseries, red] at (-7, -1.2) {ONLINE};

			% ==========================================
			% OFFLINE PART
			% ==========================================
			\node[database] (db_unlabeled) at (1, 6.5) {Unlabeled\\Dataset};

			\node[block] (data_aug_block) at (-4, 6.5) {Data\\Augmentation};

			\node[database] (db_augmented_unlabeled) at (-4, 3.5) {Augmented\\Dataset};

			\node[block] (teacher_model) at (1, 3.5) {
				\textbf{Teacher}\\
				\scriptsize Model
			};

			\node[database] (db_augmented_labeled) at (6, 3.5) {
				Augmented Labeled\\Dataset
			};

			\node[block] (student_offline) at (1, 1.5) {
				\textbf{Student}\\
				\scriptsize Model (Offline)
			};

			% Connections OFFLINE
			\draw[line] (db_unlabeled) -- (data_aug_block);
			\draw[line] (data_aug_block) -- (db_augmented_unlabeled);
			\draw[line] (db_augmented_unlabeled) -- node[above, font=\small] {Label} (teacher_model);
			\draw[line] (db_unlabeled) -- node[right, font=\footnotesize] {Fit (If needed)} (teacher_model);
			\draw[line] (teacher_model) -- (db_augmented_labeled);

			% Distillation / pretraining
			\draw[line] (db_augmented_labeled.south) |-
			node[pos=0.5, below] {Offline pre-training} (student_offline.east);

			% ==========================================
			% ONLINE PART
			% ==========================================
			\node[block] (student_online) at (1, -2.5) {
				\textbf{Student}\\
				\scriptsize Model (Online)
			};

			\node[entity] (robot) at (-4, -3.5) {\Large Robot};

			\node[decision] (uncertainty) at (3, -5.4) {
				High\\Uncertainty ?
			};

			\node[decision] (anomaly) at (-1, -5.4) {
				High\\Anomaly\\Score ?
			};

			\node[entity] (expert) at (1, -8.0) {\Large Operator};

			% ==========================================
			% TRANSITION
			% ==========================================
			\draw[draw=black, line width=0.5mm, dashed, ->]
			(student_offline) -- node[right, font=\bfseries\small] {Onboard} (student_online);

			% ==========================================
			% ONLINE FLOW
			% ==========================================
			\draw[line] (robot.north) |- node[pos=0.5, above, font=\footnotesize] {Data} (student_online.west);

			% Tronc commun (1er angle)
			\draw[thick] (student_online.south) -- (1, -3.5);

			% Branches avec 2ème angle (-|)
			\draw[line] (1, -3.5) -| (uncertainty.north);
			\draw[line] (1, -3.5) -| (anomaly.north);

			% Flèches vers l'expert (2 angles droits, sans se croiser à la fin)
			% "YES" à gauche pour l'anomalie, à droite pour l'incertitude
			\draw[line] (anomaly.south) -- node[left, font=\small\bfseries, text=red] {YES} (-1, -7.0) -| ([xshift=-0.8cm]expert.north);

			\draw[line] (uncertainty.south) -- node[right, font=\small\bfseries, text=red] {YES} (3, -7.0) -| ([xshift=0.8cm]expert.north);

			% FEEDBACK LOOPS
			% ==========================================
			\draw[redline] (expert.west) -| node[pos=0.25, above, red] {Repair} (robot.south);

			\draw[redline] (expert.east) -- node[above, red] {Adaptation} (6, -8.0) |- (student_online.east);

		\end{tikzpicture}
	}
	\caption{
		Overall architecture of the proposed framework.
	}
	\label{fig:architecture}
\end{figure}

Despite these advantages, implementing data-driven FD introduces significant challenges. Because anomalous data is inherently scarce and prohibitive to label exhaustively, fault detection systems must rely on unsupervised learning to establish a baseline of nominal operation~\cite{liu2025tsb}. Recently, high-capacity deep neural networks---such as large-scale Transformers and deep generative architectures---have demonstrated strong performance in extracting complex temporal dependencies~\cite{park2018multimodal}. However, a bottleneck prevents their direct deployment in real-world mobile robotics: the prohibitive computational cost of continuous learning~\cite{chen2019deep}. For mobile robots navigating dynamic environments, continuous learning is mandatory~\cite{lesort2020continual}. Wear and tear, payload changes, and environmental shifts dictate that a static model will inevitably trigger false positives when encountering novel but normal physical states~\cite{8496795}. Modifying a deep neural network on the edge to adapt to these drifts requires backpropagation, which is computationally expensive, memory-intensive, and inherently prone to catastrophic forgetting~\cite{parisi2019continual, hadsell2020embracing}.

To bridge the gap between the representational power of deep learning and the constraints of edge robotics, this paper proposes a Teacher-Student distillation framework for unsupervised and online anomaly detection, as illustrated in Figure~\ref{fig:architecture}. Rather than deploying computationally heavy neural networks directly on the robot, we leverage them offline as high-capacity "Teachers" to evaluate historical data and generate pseudo-labels. A lightweight, CPU-bound "Student" is then trained to approximate this complex decision boundary. Crucially, this decoupled architecture enables the Student to perform real-time weight updates onboard without the prohibitive cost of deep backpropagation. This adaptation is driven by sparse, uncertainty-triggered feedback from a human operator, who can efficiently confirm the fault state by relying on accessible cues, such as direct visual observation or external camera feeds. Ultimately, while the Teacher provides a robust statistical baseline, the Student continuously adapts to the evolving physical reality of the specific hardware.

The main contributions of this work are as follows:
\begin{itemize}
	\item An offline-to-online knowledge distillation pipeline in an unsupervised context that bypasses the backpropagation bottleneck, allowing the representational power of heavy deep learning models to be deployed on computationally constrained robotic edge CPUs.
	\item An adaptation of the MiniRocket architecture, replacing its static head with a Recursive Least Squares estimator. This enables ultra-low latency inference, uncertainty quantification, and computationally efficient online continuous learning.
	\item The integration of an uncertainty-guided active learning loop that acts as a regularizer in continuous, imbalanced data streams, preventing catastrophic forgetting caused by the over-sampling of nominal states during online adaptation.
\end{itemize}

The remainder of this paper is structured as follows: Section~\ref{sec:sota} reviews related work in data-driven FD, unsupervised time series anomaly detection, and extrinsic time series regression. Section~\ref{sec:methodo} details the proposed framework, while Section~\ref{sec:systemimpl} describes the specific system implementation for our application. Section~\ref{sec:expe} presents the experimental setup and results, evaluating the approach on both a public benchmark dataset and a real-world robotic scenario. Finally, Section~\ref{sec:conclu} concludes the paper and outlines directions for future research.

\section{Background}
\label{sec:sota}

\subsection{Data-Driven Fault Detection}

Data-driven FD constructs statistical or learned representations of nominal robotic behavior to identify operational deviations. Because physical hardware faults are inherently rare and structurally diverse, data-driven FD is predominantly formulated as an anomaly detection problem~\cite{park2018multimodal, willibald2025multimodal, chen2020unsupervised}. While these methodologies demonstrate strong offline detection accuracy, they often struggle with the physical realities of mobile robotics: the necessity for ultra-low-latency execution and the requirement for continuous online adaptability in the face of mechanical wear.

% Recently, anomaly detection using exteroceptive visual data has seen significant advancements~\cite{lanighan2024online, jiang2025anomalies, jin2023anomaly}. While vision-based methods are powerful, they introduce substantial computational overhead and latency. In this work, we restrict our study to internal proprioceptive and kinematic sensors. Anomalies in internal sensor data inherently capture a vast majority of physical faults (e.g., motor degradation, actuator failures, mechanical wear) directly at the source, offering a lower-dimensional, highly responsive, and computationally efficient alternative to high-bandwidth visual streams.

To address the computational constraints of robots, recent literature has pivoted toward lightweight and online approaches. For instance,~\cite{katta2023towards} demonstrated a feature-selection and sliding-window framework for reliable, low-cost onboard fault detection in UAVs. Similarly, the approach in~\cite{boelter2025model} applied time-series subspace analysis directly to drill avionics telemetry for unsupervised anomaly detection, avoiding the computational overhead of exteroceptive sensors. Other online methodologies leverage world models in reinforcement learning to continuously monitor discrepancies between predicted and observed system behaviors~\cite{domberg2025world}, or employ data-driven Koopman operators to act as real-time digital twins for mechanical components~\cite{pumphrey2025data}. Despite achieving real-time inference, these architectures rely on models that remain entirely static after offline training, lacking the continuous learning capabilities necessary to adapt to mechanical wear over a robot's deployment lifecycle.

A promising avenue for achieving both speed and adaptability is transforming a complex unsupervised anomaly detector into a faster supervised model, as initially explored in~\cite{khalastchi2017hybrid}. Our proposed framework extends this paradigm across three dimensions. First, we transition from static tabular data to high-frequency multivariate time series, essential for capturing the complex temporal dependencies of modern mobile robots. Second, to overcome the scarcity of real-world fault data, we introduce a data injection strategy to synthesize anomalous boundaries. Finally, our architecture is designed for continuous online learning, ensuring the model adapts to evolving hardware dynamics over time.

\subsection{Unsupervised Time Series Anomaly Detection}

Unsupervised Time Series Anomaly Detection (TSAD) aims to identify temporal observations that deviate significantly from an expected nominal baseline without relying on labeled training data. In time series data (like in a robotic context), anomalies do not manifest only as isolated point anomalies (e.g., sensor spikes) but also as complex subsequence anomalies (e.g., mechanical drift), necessitating specialized detectors capable of capturing temporal dependencies~\cite{boniol2024dive}.

While a multitude of methods have been explored, recent advancements in deep learning have significantly pushed the state-of-the-art in multivariate TSAD~\cite{liu2024elephant,boniol2024dive}. Most notably, time-series foundation models have demonstrated powerful zero-shot detection capabilities~\cite{ekambaram2026tspulse}. By leveraging massive pre-training datasets, these architectures can establish robust boundaries of normality. However, despite their unprecedented accuracy, their immense parameter counts and rigid contextual windows render them incompatible with the ultra-low latency and computational constraints of edge robotic deployment.

\subsection{Extrinsic Time Series Regression}

Extrinsic Time Series Regression (ETSR) refers to the task of mapping an input time series to a continuous score variable~\cite{mohammadi2024deep}. Among the algorithms, MiniRocket has established itself as a robust architecture, achieving state-of-the-art accuracy with exceptional computational efficiency~\cite{dempster2021minirocket,middlehurst2024bake}. MiniRocket extracts temporal features by convolving the input sequence with a predefined, deterministic set of fixed kernels across various size. The features generated by the proportion of positive values pooling are then evaluated by a ridge regressor.

While architectures like MiniRocket excel in static ETSR, the literature currently lacks lightweight, natively adaptive algorithms capable of continuous online learning. Existing streaming regression models are primarily designed for tabular data and struggle to natively extract temporal features. Conversely, adaptive deep time-series models require computationally prohibitive backpropagation on the edge~\cite{chen2019deep,parisi2019continual}. To bridge this gap, we modified the architecture by separating the fixed temporal feature extractor from an adaptive learning head, making online ETSR practical for robotics.

\section{Overall Framework}
\label{sec:methodo}

This section details the proposed framework and its methodological components. As illustrated in Figure~\ref{fig:architecture}, our architecture operates on an offline-to-online knowledge distillation pipeline. The framework aims to train a computationally lightweight, supervised ETSR model (the Student) using the anomaly scores generated by a complex, unsupervised TSAD model (the Teacher). Real-world robotic data consists of a majority of nominal instances with a minority of anomalies. To mitigate this imbalance, the offline pipeline first injects anomalies via data augmentation. The Teacher subsequently evaluates these augmented sequences, generating pseudo-labels to establish robust decision boundaries for the Student. Once onboarded, the Student executes high-frequency tracking and dynamically adapts to human operator feedback via an event-triggered loop.

\subsection{Problem Setup}

Let $\mathcal{S} \in \mathbb{R}^{T \times C}$ denote a continuous, multivariate time-series stream acquired from a robotic system, consisting of $T$ consecutive observations across $C$ sensors. To facilitate sequential processing and temporal feature extraction, this continuous sequence is partitioned into a series of $N$ overlapping sliding windows. Let $\mathcal{D}_{train} = \{X_1, X_2, \dots, X_N\}$ represent the resulting unlabeled training dataset, where each instance $X_i \in \mathbb{R}^{W \times C}$ is a temporal window of fixed length $W$. The objective of TSAD is to learn a mapping function $f: \mathbb{R}^{W \times C} \rightarrow \mathbb{R}$ that assigns a continuous anomaly score $s_i$ to each window $X_i$. A high score indicates a statistical or dynamic deviation from the learned nominal behavior, signifying a potential fault. Because $\mathcal{S}$ is collected during standard operations, $\mathcal{D}_{train}$ is assumed to consist overwhelmingly of nominal data even if some anomalies may be present.

\subsection{Teacher-Student Pipeline}

To bypass the need for explicit human labeling while ensuring high performance, we formalize our framework as a Teacher-Student knowledge distillation pipeline. In this paradigm, the Teacher, denoted as $f_T$, evaluates the unlabeled dataset offline to generate pseudo-labels $y_i = f_T(X_i)\in \mathbb{R}$.

Although mechanical degradation in physical systems is inherently a continuous process, the raw anomaly scores generated by the Teacher can exhibit high-frequency volatility due to sensor noise and windowing artifacts. To align the pseudo-labels with physical reality, we apply a rolling moving average filter to extract the true underlying degradation trend. Subsequently, these smoothed scores are standardized (z-score normalized) to ensure numerical stability, scale-invariant weight updates, and easier convergence for the Student's linear regression head.

Let $\hat{y}_i$ denote this filtered and standardized target. To prioritize the accurate detection of severe physical faults over minor fluctuations, we introduce a weighting mechanism during the offline training phase. Specifically, we assign an exponential sample weight $\gamma_i = \exp(\hat{y}_i - \min(\hat{y}))$ to each temporal window. By subtracting the global minimum, nominal data receives a baseline weight near $1$, while highly anomalous instances receive an exponential multiplier. The refined pseudo-labeled dataset is thus formalized as $\mathcal{D}_{labeled} = \{(X_i, \hat{y}_i, \gamma_i)\}_{i=1}^N$. The Student model, denoted as $f_S$, is designed as a continuous learning regressor whose objective is to approximate the Teacher's behavior by minimizing the weighted regression error across $\mathcal{D}_{labeled}$.

\subsection{Data Augmentation}

Training an ETSR model requires mitigating the class imbalance inherent to anomaly detection. If the Student regressor is trained exclusively on normal data, its weights will collapse toward the normality baseline. Consequently, it will fail to accurately regress high anomaly scores during inference, leading to an unacceptable rate of false negatives.

To address this temporal scarcity and facilitate knowledge transfer, we use a data injection strategy. We generate $n_{aug}$ augmented sequences based on the original dataset by injecting synthetic anomalies of random lengths, at random positions, and across random features. To simulate diverse hardware and environmental faults, the injected patterns are randomly sampled from four categories: spikes, cutoffs, drifts, and noise. Crucially, these augmented sequences are subsequently processed by the unsupervised Teacher model to extract their expected anomaly scores and weights, yielding the augmented dataset $\mathcal{D}_{aug}$. The final training corpus for the Student regressor is the union of the original and augmented datasets, denoted as $\mathcal{D}^*_{labeled} = \mathcal{D}_{labeled} \cup \mathcal{D}_{aug}$.

\subsection{Online Execution and Continuous Adaptation}

Once the Student has been trained from the new pseudo-labeled dataset $\mathcal{D}^*_{labeled}$, it is deployed directly onto the robot's embedded hardware. During active operation, the system sequentially evaluates streaming time series to output real-time anomaly scores. To preserve the integrity of the high-frequency control loop, this forward inference step must satisfy hard real-time latency constraints. Furthermore, the Student should be capable of continuous learning.

Because the Student is trained to output a normalized anomaly score, the feedback mechanism of the continuous learning accepts both discrete targets (e.g., 0 for a nominal state, 1 for an anomalous state) and---given the regression nature of the task---continuous intermediate values to represent partial degradation or state uncertainty. To avoid a computational bottleneck, the online learning of the Student model must also be executed with minimal latency.

\section{System Implementation}
\label{sec:systemimpl}

While our proposed framework is inherently model-agnostic and compatible with diverse TSAD and ETSR models, this section details the system implementation and the specific Teacher and Student models selected for our experimental validation.

\subsection{The Teacher Model}

In our distillation framework, the Teacher model acts as an offline Oracle. Freed from the strict latency and hardware constraints of the robotic edge, the Teacher can leverage massive, high-capacity architectures. While this could theoretically be instantiated by any deep neural network pre-trained on extensive domain-specific robotic data, Foundation Models offer a superior alternative by eliminating the need for dataset-specific fine-tuning, model evaluation, and selection.

For this implementation, we selected TSPulse~\cite{ekambaram2026tspulse} as the Teacher. As a leading time-series foundation model, TSPulse demonstrates highly robust zero-shot generalization capabilities~\cite{liu2024elephant}. Its primary advantage is its strong zero-shot generalization. Architecturally, TSPulse enforces a rigid temporal context window of $L=512$ time steps.

\subsection{The Embedded Student}

To satisfy the stringent computational constraints of the robotic edge, the Student must be fast, light, and adaptive. We selected MiniRocket~\cite{dempster2021minirocket} as our base CPU-bound regressor. Mathematically, it applies a fixed set of convolutional kernels to an incoming time-series window $X_t$, extracting a high-dimensional feature vector $x_t = \Phi(X_t) \in \mathbb{R}^k$ via the proportion of positive values.

To enable continuous online adaptation without backpropagation, we replace MiniRocket's default static ridge head with a Recursive Least Squares (RLS) estimator. For a target pseudo-label $y_t$ and a forgetting factor $\lambda \in (0, 1]$, the decision weights $w_t$ and inverse covariance matrix $P_t$ are updated sequentially:
$$e_t = y_t - x_t^T w_{t-1}, \quad g_t = \frac{P_{t-1} x_t}{\lambda + x_t^T P_{t-1} x_t}$$
$$w_t = w_{t-1} + g_t e_t, \quad P_t = \frac{1}{\lambda} (P_{t-1} - g_t x_t^T P_{t-1})$$
This isolates all training overhead to the linear weights ($w_t$), keeping the heavy feature extraction ($\Phi$) frozen.

Crucially, the RLS inherently quantifies predictive uncertainty. This absolute uncertainty relies on the geometric novelty in the feature space, $u_t = x_t^T P_{t-1} x_t$---which is conveniently pre-computed in the denominator of $g_t$---scaled by a dynamic noise variance $\sigma^2_t$. We track $\sigma^2_t$ via an exponentially weighted moving average of the \textit{a posteriori} error. To minimize computational cost, we trigger the $O(k^2)$ matrix update ($P_t$) and operator feedback requests only when $\sigma^2_t u_t \ge \tau$. Otherwise, the model state is frozen.

\section{Experimentation}
\label{sec:expe}

After detailing the experimental setup in Section~\ref{sec:setup}, we divide our evaluation into two phases. First, in Section~\ref{sec:generalapp}, we evaluate the framework's foundational properties. Specifically, we quantify the impact of data injection and benchmark the student's inference and adaptation speeds. Second, in Section~\ref{sec:robotapp}, we deploy the complete architecture on a mobile robotics dataset to evaluate online adaptation and resilience to catastrophic forgetting in physical environments.

\subsection{Experimental Setup}
\label{sec:setup}

We used the pre-trained TSPulse foundation model as the offline oracle. For the Student model, we deployed the MiniRocket implementation from \texttt{aeon}~\cite{middlehurst2024aeon}, which leverages Numba and NumPy for CPU-optimized execution. After the offline training, we replace the static ridge regressor by our RLS estimator with a forgetting factor of $\lambda = 0.99$.

We evaluate performance using the Volume Under the Surface of the Precision-Recall curve (VUS-PR)~\cite{boniol2025vus}. Unlike standard point-wise metrics, VUS-PR is explicitly designed for range-based time series; it penalizes late detections and prolonged false alarms while tolerating minor temporal shifts, accurately reflecting robotic deployment realities. To ensure reproducibility, all random seeds are fixed and the source code is available on GitHub\footnote{https://anonymous.4open.science/r/ICRA2027-AB08}.

\subsection{Foundational properties of the framework}
\label{sec:generalapp}

Before evaluating the continuous online adaptation in real-world scenarios, we first establish the foundational properties of the distilled framework. This section evaluates two criteria: first, how accurately the static Student regressor replicates the Teacher's decision boundary to provide a robust "Warm Start" initialization; and second, whether the Student's inference and RLS update mechanisms satisfy the latency and memory constraints required for onboard robotic deployment.

\subsubsection{Evaluation of the Data Augmentation}
\label{sec:eval_aug}

\begin{figure}[t!]
	\centering
	\includegraphics[width=\linewidth]{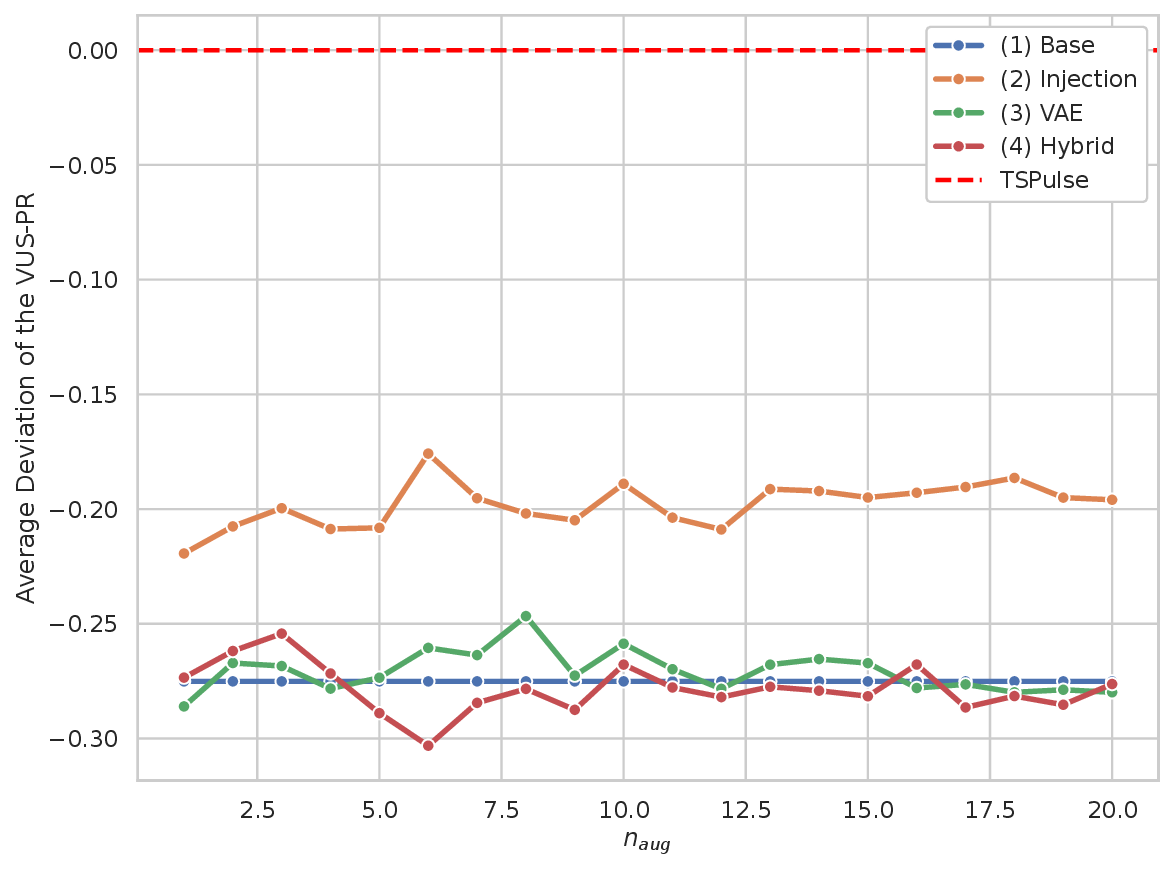}
	\caption{Global ablation study comparing the VUS-PR performance of TSPulse and MiniRocket across different data augmentation strategies.}
	\label{fig:data_aug}
\end{figure}

We use the comprehensive TSB-AD benchmark~\cite{liu2024elephant} to validate the foundational mechanics of our distillation pipeline. While TSB-AD encompasses multiple application domains, its advantage lies in the massive structural diversity of its temporal anomalies, ranging from isolated point spikes to complex, long-term contextual shifts. This variety allows us to rigorously stress-test the robustness of our data augmentation strategies against diverse fault signatures.

We conduct an ablation study to evaluate the role of data augmentation in the distillation pipeline. We establish the Teacher as the performance upper bound and compare it against the distilled Student trained under four distinct data augmentation regimes: (1) a baseline with zero augmentation, (2) our proposed injection method, (3) Variational Autoencoder (VAE) generation, and (4) a hybrid approach combining injection and VAE. Furthermore, we vary the augmentation factor ($n_{aug}$) to quantify the impact of the number of augmented datasets on the Student's performance.

To rigorously isolate and evaluate the efficiency of our data augmentation, it is imperative to eliminate the confounding variable of Teacher inaccuracy. The Student model cannot reconstruct decision boundaries that the Teacher fails to identify. Consequently, from TSB-AD, we selected datasets where TSPulse has more than $0.30$ VUS-PR (i.e., TAO, Exathlon, LTDB, SVDB, SMD, Daphnet, SMAP). This filtering ensures that our ablation study strictly measures the Student's ability to approximate a competent Teacher.

Figure~\ref{fig:data_aug} presents the aggregated results across all dataset families. The necessity of data augmentation is apparent when comparing the baseline Student to the Teacher: without augmentation, MiniRocket yields a VUS-PR score averaging 0.27 below TSPulse. The primary objective of the data augmentation is to close this generalization gap.

Comparing the augmentation strategies, empirical results show that the injection technique consistently outperforms both the baseline Student and the alternative augmentation methods. Conversely, the VAE and Hybrid techniques fail to meaningfully improve the Student's performance. While not definitively proven in this study, we hypothesize that the VAE struggles to synthesize physically compliant kinematic anomalies, potentially generating out-of-distribution artifacts. Under this assumption, when the Teacher evaluates these nonphysical artifacts, it would likely propagate erroneous confidence scores, ultimately corrupting the Student's learned decision boundary.

Regarding the scaling parameter $n_{aug}$ within the injection technique, results indicate a performance increase up to $n_{aug} = 6$. However, for $n_{aug} > 6$, the Student's performance plateaus around a $-0.2$ VUS-PR deficit relative to the Teacher. This stagnation is fundamentally tied to the Student's architectural capacity constraint. Because MiniRocket relies on a fixed set of convolutional kernels paired with a linear regression head, its feature space eventually saturates. Beyond $n_{aug} = 6$, additional augmented samples provide highly redundant information that falls within already established linear decision boundaries.

\subsubsection{Computational Efficiency and Edge Deployability}

While the Teacher-Student distillation framework aims at preserving the Teacher detection capabilities, its primary objective is to satisfy the stringent Size, Weight, and Power (SWaP) constraints of mobile robotic platforms. To empirically validate the deployability of our Student, we evaluated its computational footprint on a standard embedded edge computer: an NVIDIA Jetson Orin Nano Super. To make the Student deployable online, two criteria are required. First, the Student should be able to make detections with low latency while maintaining a low memory footprint. Second, the Student should be able to learn online with a low computational cost.

\paragraph{Evaluation of the cost of online inference}

\begin{table*}[htb!]
	\centering
	\caption{Hardware Benchmark on NVIDIA Jetson Orin Nano (Mean over 10 runs)}
	\label{tab:hardware_bench}
	\begin{tabular}{llcccc}
		\hline
		\textbf{Model}       & \textbf{Target Hardware} & \textbf{Context Window ($L$)} & \textbf{Mean Latency}     & \textbf{99th Percentile}  & \textbf{Peak OS Memory}     \\
		\hline
		Teacher (TSPulse)    & GPU                      & 512                           & $36.09\pm0.95$ ms         & $54.73\pm1.74$ ms         & $804.87\pm7.28$ MB          \\
		Teacher (TSPulse)    & CPU                      & 512                           & $343.45\pm6.26$ ms        & $530.64\pm14.94$ ms       & $555.54\pm5.72$ MB          \\
		\hline
		Student (MiniRocket) & CPU                      & 512                           & $25.00\pm0.21$ ms         & $25.97\pm0.35$ ms         & $546.78\pm2.25$ MB          \\
		Student (MiniRocket) & CPU                      & 100                           & \textbf{4.30$\pm$0.02 ms} & \textbf{4.50$\pm$0.09 ms} & \textbf{546.32$\pm$2.76 MB} \\
		\hline
	\end{tabular}
\end{table*}

When evaluating edge viability, the size of the model weights is often overshadowed by the execution environment. As Table \ref{tab:hardware_bench} demonstrates, the Peak OS Memory for both TSPulse and MiniRocket on the CPU converges around $\sim$550 MB. This baseline represents the inescapable framework footprint required to run Python, PyTorch, or Scikit-Learn/Numba on the Jetson's unified memory. Consequently, deploying the Teacher model on the CPU offers no memory advantage over the Student. Furthermore, attempting to accelerate the Teacher model using the Jetson's GPU triggers the initialization of the CUDA context, imposing a memory penalty. The total process footprint inflates to $804.87$ MB, which consumes even more memory simply to enable inference.

Given the identical CPU memory budget ($\approx$546-555~MB), the evaluation shifts to computational efficiency. The Teacher fails to deliver real-time performance on the ARM CPU, exhibiting a mean latency of $343.45$~ms and severe jitter. Conversely, the Student model operating on the same CPU with an identical context window ($L=512$) achieves a mean latency of $25.00$ ms. While reducing the context window might theoretically complicate the detection of prolonged sequence anomalies, empirical results on our filtered TSB-AD benchmark demonstrate that a smaller context window actually improves the Student's precision. With the injection data augmentation, optimizing the window to $L=100$ yields an average VUS-PR improvement of $+0.14$ over the $L=512$ baseline. Consequently, deploying this optimized $L=100$ configuration not only maximizes detection performance but simultaneously drives latency down to a minimal $4.30$~ms. Crucially for hard real-time robotic applications, MiniRocket demonstrates determinism. The tight bounds between its mean and 99th percentile latency ($4.30$~ms vs. $4.50$~ms) guarantee highly predictable execution cycles, free from the heavy OS-level variance observed in the PyTorch/GPU pipeline.

\paragraph{Evaluation of the cost of online learning}

\begin{table*}[htb!]
	\centering
	\caption{Computation time of the update of the RLS depending on the number of kernels $k$ of MiniRocket on NVIDIA Jetson Orin Nano (Mean over 10 runs)}
	\label{tab:exp_time}
	\begin{tabular}{lccccccc}
		\hline
		$k$       & 100           & 500           & 1,000         & 2,000          & 4,000          & 8,000           & 10,000          \\
		\hline
		Time (ms) & $0.06\pm0.01$ & $0.79\pm0.01$ & $3.05\pm0.06$ & $12.08\pm0.15$ & $62.00\pm0.05$ & $242.70\pm0.09$ & $330.27\pm0.14$ \\
		\hline
	\end{tabular}
\end{table*}

While achieving low inference latency is crucial, the computational cost of the online learning phase must also be considered. MiniRocket uses $k$ random convolutional kernels to extract features, which are then passed to an RLS estimator. A key issue with RLS is the $\mathcal{O}(k^2)$ quadratic complexity required to update the covariance matrix; using more kernels drastically increases the update time. While higher values of $k$ yield better accuracy~\cite{dempster2021minirocket}, reducing $k$ down to $2{,}000$ leads to a negligible performance drop on standard benchmarks. We measured the update time on the Jetson across different values of $k$ (Table~\ref{tab:exp_time}). While the default setting of $k=10{,}000$ kernels results in an update latency of $330.27$ ms, using $k=2{,}000$ drops the update time to under $15$ ms, easily meeting the real-time constraints of our robotic application.

\subsection{Real-World Robotic Application}
\label{sec:robotapp}

While standard benchmarks validate the general capabilities of our framework, real-world robotics introduces complex physical and environmental constraints. To evaluate our system in the field, we deployed it on a four-wheeled mobile robot. In this section, we first detail our dataset and then evaluate the online adaptation of the Student model during active navigation.

\subsubsection{Dataset Generation}

We collected a real-world dataset on a university campus under uncontrolled, dynamic conditions. The robot autonomously executed point-to-point navigation missions in public spaces, continuously interacting with pedestrians and the evolving environment.

To construct the training set, we recorded 7 missions ranging from 2 to 11 minutes in duration. The dataset contains 63 continuous channels---including proprioceptive wheel status, battery consumption, and inertial sensors---sampled at a rate of 50 Hz. Although visually supervised, the data may contain hidden anomalies. Nevertheless, our unsupervised Teacher establishes a nominal baseline from these uncurated logs without manual cleaning.

\begin{figure}[t!]
	\centering
	\includegraphics[width=0.32\linewidth]{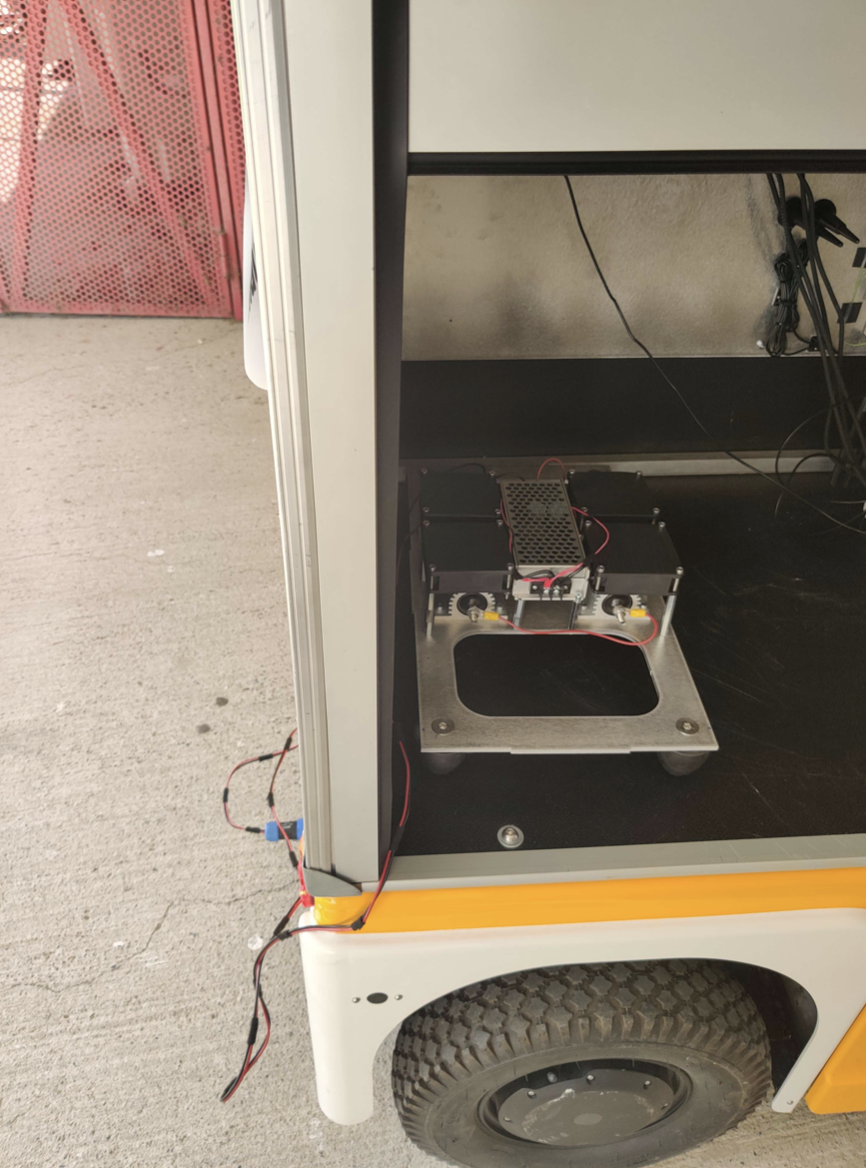}\hfill
	\includegraphics[width=0.32\linewidth]{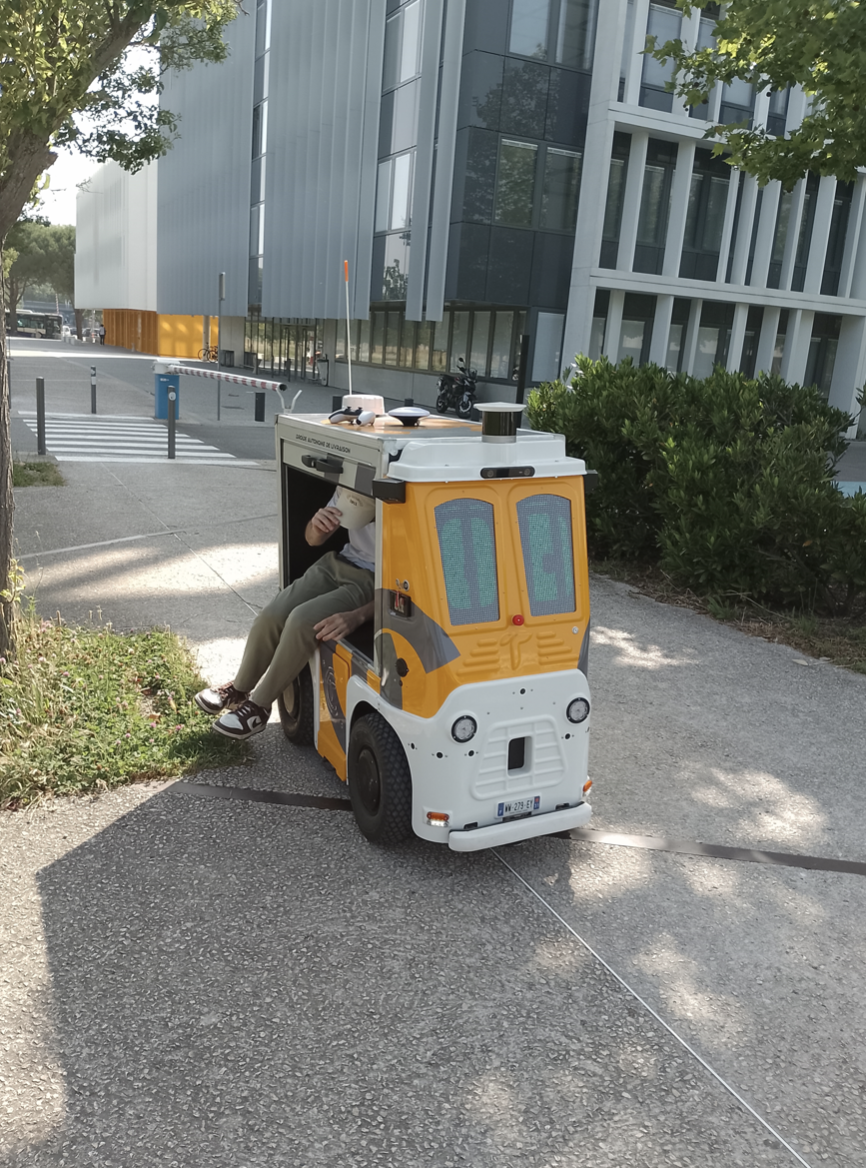}\hfill
	\includegraphics[width=0.32\linewidth]{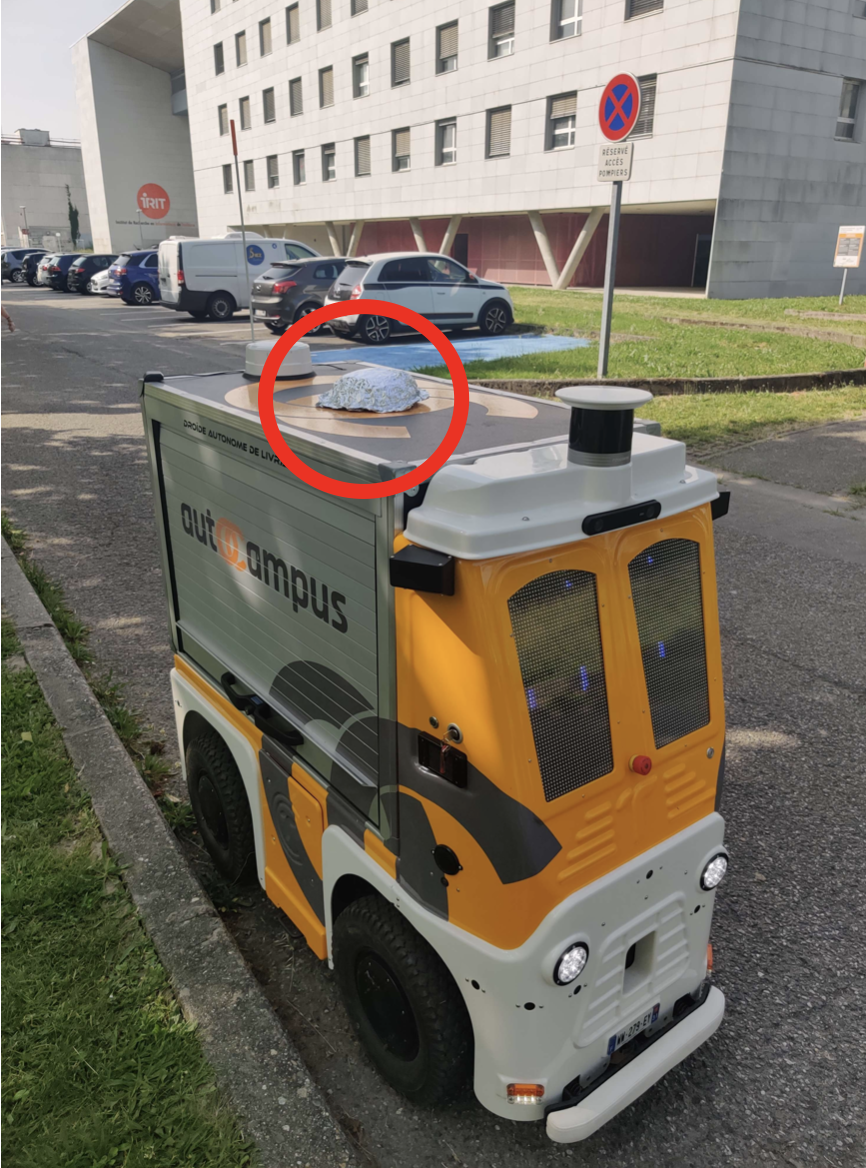}

	\caption{Pictures of induced faults. Left: Excessive power consumption. Middle: Overweight. Right: GNSS jamming.}
	\label{fig:faults}
\end{figure}

To construct the evaluation test set, we physically induced 3 distinct faults, as illustrated in Figure~\ref{fig:faults}.

\begin{itemize}
	\item \textbf{Excessive power consumption}: A module containing electrical resistors was plugged to the robot to simulate excessive power consumption.
	\item \textbf{Overweight}: A human got inside the robot to overload it and destabilise it during a mission.
	\item \textbf{GNSS jamming}: A sheet of aluminium foil was placed over the GNSS sensor to simulate a poor-quality signal.
\end{itemize}

To ensure evaluation diversity, each fault type was recorded across two separate navigation missions. Because these faults were intentionally induced, we established precise ground-truth labels for the test set by annotating the exact temporal windows.

\subsubsection{Online Learning}

\begin{figure*}[htbp]
	\centering
	\includegraphics[width=\linewidth]{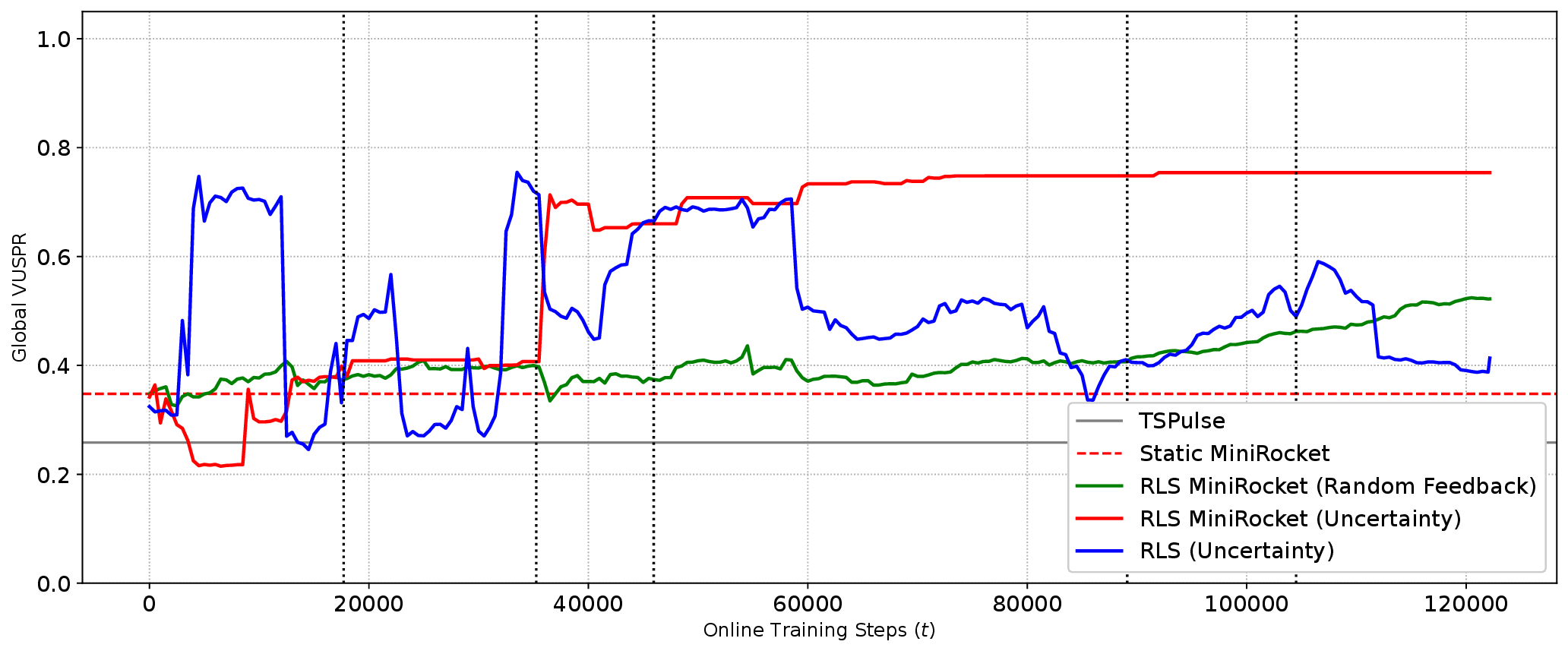}
	\caption{Evolution of the cumulative VUS-PR performance over time. Vertical lines indicate the transitions between distinct physical fault datasets.}
	\label{fig:rocket_evolution}
\end{figure*}

\begin{figure}[htbp]
	\centering
	\includegraphics[width=\linewidth]{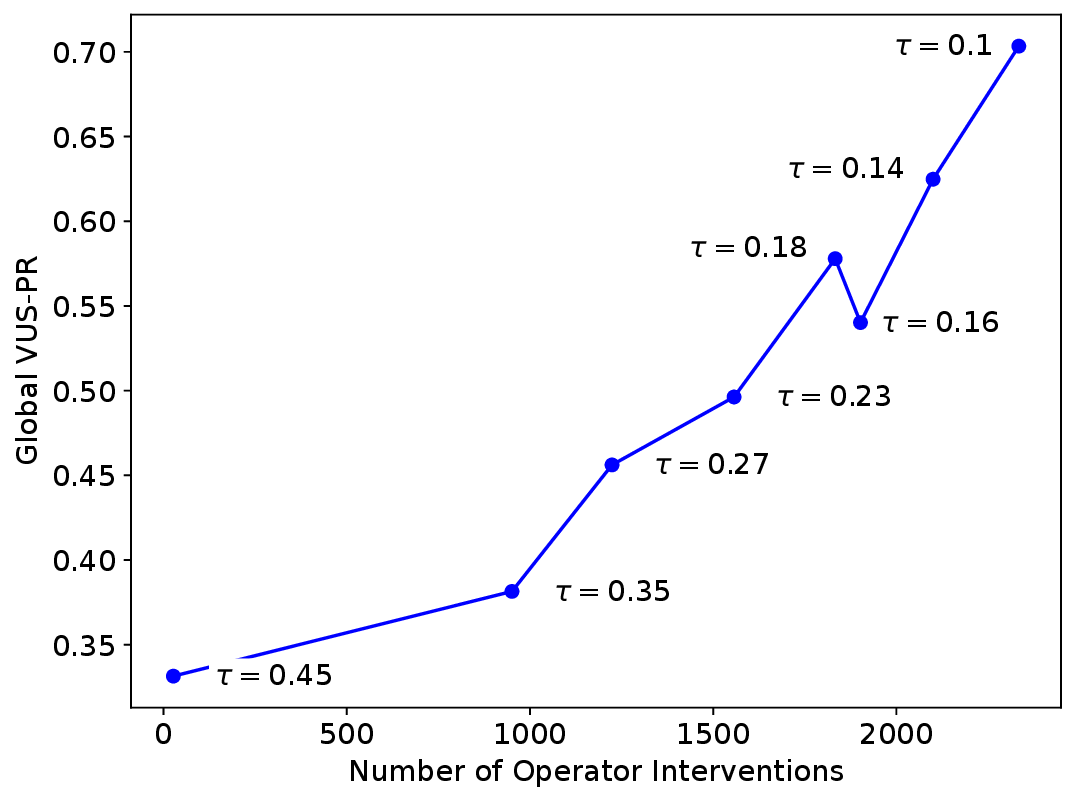}
	\caption{Pareto front showing the trade-off between detection performance (Global VUS-PR) and operator cognitive load (number of interventions) across various uncertainty thresholds ($\tau$).}
	\label{fig:pareto}
\end{figure}

To evaluate the online adaptation capabilities of the student model, we initially trained the adaptive MiniRocket using the anomaly scores generated by the TSPulse oracle. Following the optimal configuration identified in the previous section, we generated $n_{aug}~=~6$ synthetic datasets containing injected faults and trained MiniRocket with $k=2000$ kernels. After the offline distillation phase, we simulated a continuous real-time deployment by sequentially replaying the physical fault datasets.

To isolate the contributions of our architectural choices, we benchmarked our framework against multiple baselines. First, to validate the necessity of temporal feature extraction, we compared our adaptive MiniRocket against a naive RLS applied at each time step of the time series. Second, to evaluate operational feasibility, we contrasted two active learning strategies: a naive random sampling approach (where the model receives ground-truth feedback with a fixed $1\%$ probability) and our proposed uncertainty-guided strategy (where feedback is only requested when the predictive uncertainty exceeds a threshold of $\tau = 0.01$). We compute the VUS-PR of the models every 500 time steps to monitor both the acquisition of novel fault signatures and the resistance to catastrophic forgetting.

Figure~\ref{fig:rocket_evolution} reveals a contrast between static and adaptive paradigms, while also highlighting the practical limits of naive continuous learning strategies. The offline Teacher and static Student baselines fail to generalize to novel conditions, stagnating at a VUS-PR of approximately $0.26$ and $0.35$, respectively. In contrast, the uncertainty-driven adaptive MiniRocket successfully mitigates domain shifts, climbing to a VUS-PR of $0.75$. Crucially, our results demonstrate that MiniRocket's feature extraction acts as a regularizer against destructive weight updates. While the raw RLS initially shows rapid adaptive capabilities, it suffers from severe catastrophic forgetting with volatility before collapsing to a VUS-PR of $0.41$, proving that high-dimensional convolutional feature spaces are required to stabilize continuous learning. Furthermore, while relying on a random feedback strategy does yield steady improvement over the static baseline (reaching a VUS-PR of $0.52$), it remains highly inefficient. Executing this random strategy requires approximately 12,000 human interventions across the sequence. This represents an unsustainable cognitive load for human operators, highlighting the necessity of uncertainty-based sampling to maximize predictive performance while strictly minimizing the labeling budget.

To evaluate the operational viability of the active learning framework, we analyzed the sensitivity of the system to the uncertainty threshold ($\tau$). Figure~\ref{fig:pareto} illustrates the Pareto front between the final detection performance and the number of operator interventions required across the deployment. The results demonstrate a trade-off between model accuracy and operator cognitive load. While lower thresholds generally yield higher performance, the progression exhibits local instabilities. Notably, lowering the threshold from $\tau=0.18$ to $\tau=0.16$ paradoxically degrades the VUS-PR from $0.58$ to $0.54$, despite increasing operator interventions from $1833$ to $1902$. We infer that at this specific boundary, hyper-frequent weight updates expose the RLS estimator to an overwhelming majority of nominal states, inducing a class imbalance that makes the model overly conservative and degrades anomaly recall. Consequently, $\tau \in [0.18, 0.27]$ represents the optimal operational zone, reducing human interventions from the peak of $2334$ down to the $1224\text{-}1833$ range while maintaining stable detection accuracy. Finally, empirical observations of the online learning reveal that the model's uncertainty naturally decays over time as it successfully maps novel fault distributions. Consequently, the operational lifecycle follows a "warm-up" paradigm: the system requires a multitude of operator feedback upon initially encountering a domain shift, but progressively regains autonomy. This temporal decrease in required oversight is necessary for long-term, real-world robotic deployments.

\section{Conclusion}
\label{sec:conclu}

This paper introduced a framework to distill a high-capacity, static unsupervised TSAD model into a lightweight, adaptive model for mobile robotics. Because transitioning to embedded supervised models requires labeled data, we used the offline teacher to generate pseudo-labels from the training set. To overcome class imbalance and physical fault data scarcity, we used a data injection strategy, improving the model's decision boundaries over naive baselines. We also adapted a state-of-the-art ETSR model by replacing its static readout head with an RLS estimator. This modification bridged the gap between offline deep learning and edge deployment, enabling the student model to initialize and continuously adapt online via uncertainty-driven feedback.

Despite these promising results, our framework presents three primary limitations. First, evaluating long-term resilience to catastrophic forgetting is constrained by the current lack of continuous, multi-hour robotic benchmarks featuring sequential fault transitions. Second, while the RLS uncertainty effectively triggers queries, high-dimensional temporal features remain opaque, forcing the human operator to evaluate raw sensor signals without diagnostic guidance. Finally, our synthetic data augmentation strategy relies on relatively simple statistical perturbations, which may not capture complex physical dynamics across all fault modes.

To address these limitations, future work will pursue three main directions. First, we plan to establish an open lifelong learning benchmark to systematically stress-test online adaptation under prolonged operational drifts. Second, we will investigate Explainable AI techniques to interpret the temporal features driving the model's uncertainty, thereby reducing operator cognitive load and streamlining interactive feedback. Third, we will enhance the augmentation pipeline by incorporating simulation-driven fault injection and physics-informed models to generate physically consistent anomaly signatures.

\section*{ACKNOWLEDGMENT}
The authors would like to thank Prof. Geoff Webb for his valuable insights regarding the ROCKET architecture, Eng. Adrian Hadi for his technical assistance with the Jetson and Maxime Monfraix for his help with the accompanying videos.

\bibliographystyle{plain}
\bibliography{bib}

\end{document}